\documentclass[sigconf]{acmart}

\AtBeginDocument{%
  }

\setcopyright{acmlicensed}
\copyrightyear{2026}
\acmYear{2026}
\acmDOI{XXXXXXX.XXXXXXX}
\acmConference[Conference acronym 'XX]{Make sure to enter the correct
  conference title from your rights confirmation email}{June 03--05,
  2026}{Woodstock, NY}
\acmISBN{978-1-4503-XXXX-X/2018/06}

\begin{document}

\title{Incremental Multilingual Text2Cypher with Adapter Combination}

\author{Makbule Gulcin Ozsoy}
\email{makbule.ozsoy@neo4j.com}
\affiliation{%
  \institution{Neo4j}
  \city{London}
  \country{UK}
}


\renewcommand{\shortauthors}{Ozsoy et al.}

\begin{abstract}
Large Language Models (LLMs) enable natural language interfaces for graph databases, translating user questions into structured queries via tools like Text2SQL, Text2SPARQL or Text2Cypher. While these systems improve database accessibility, most existing research focuses on English with limited multilingual support. This work proposes a multilingual Text2Cypher framework that allows incremental addition of new languages without retraining the full model, while avoiding manual hyperparameter tuning and maintaining performance close to joint multilingual fine-tuning. Our approach trains language-specific LoRA adapters for English, Spanish and Turkish and combines them via either uniform linear merging or a learned fusion MLP with dynamic adapter weighting. Experiments show that the fusion MLP recovers ~75\% of the accuracy gains of joint multilingual fine-tuning while requiring substantially fewer training examples. Additionally, low-resource languages achieve competitive performance, demonstrating both efficiency and practical applicability for multilingual Text2Cypher systems. Overall, this work demonstrates that adapter fusion provides an effective method to combine multiple language-specific models, balancing performance, data efficiency and ease of extension in multilingual graph query systems.

\end{abstract}




\maketitle

\section{Introduction}


Database query languages such as SQL (for relational databases), SPARQL (for RDF graphs) and Cypher (for graph databases) enable structured and efficient access to data. Recent advances in large language models (LLMs) have enabled natural language interfaces, such as Text2SQL, Text2SPARQL and Text2Cypher, which translate user questions into executable database queries. While these systems improve accessibility, most prior work focuses on English, leaving multilingual querying of graph databases underexplored~\cite{jannuzzi2024zero,geng2024not}.

\begin{figure}
  \centering
  \includegraphics[width=0.98\linewidth]{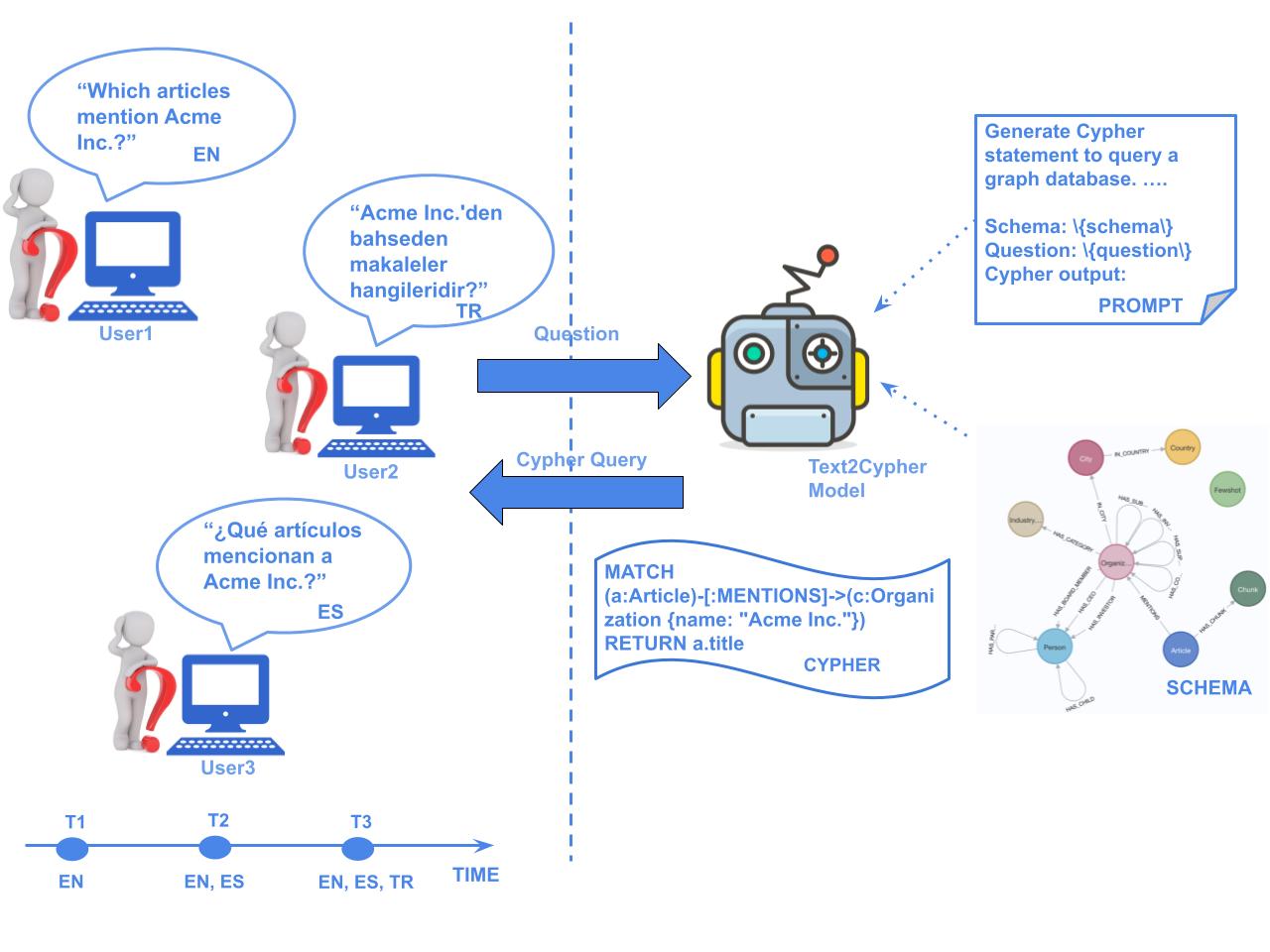}
    \caption{Incremental language expansion in multilingual Text2Cypher. At T1, only English is supported. Spanish (T2) and Turkish (T3) are added via new LoRA adapters + MLP retraining, without touching existing adapters. All speakers receive identical Cypher output.
    }
  \label{fig:text2cypher}
\end{figure}

Prior work~\cite{ozsoy2025t2cmultilang} established a benchmark across English, Spanish and Turkish, showing that base models favor the high-resource English language over medium- and low-resource languages. Although joint multilingual fine-tuning narrows this performance gap, it requires retraining the full model whenever a new language is added, which is computationally expensive. Figure~\ref{fig:text2cypher} illustrates incremental language expansion: initially (T1) only English is supported. At later stages (T2, T3), Spanish and Turkish are added. Users query the same graph database in all supported languages, expecting consistent Cypher outputs such as:
\texttt{MATCH (a:Article)-[:MENTIONS]->(c:Company {name: "Acme Inc."}) RETURN a.title}.

In this work, we address three practical challenges for multilingual Text2Cypher: (i) adding new languages incrementally without full retraining, (ii) avoiding manual hyperparameter tuning and (iii) maintaining performance close to joint multilingual fine-tuning, particularly for low-resource languages. For this purpose, we train language-specific LoRA adapters for English, Spanish and Turkish, combining them via uniform linear merging or a learned fusion MLP with dynamic adapter weighting. 
We focus on efficient, practical strategies for expanding language support, leveraging previously generated outputs and a fixed base model to concentrate on adapter fusion and incremental scaling.

Our main contributions are:
\begin{itemize}
\item We investigate adapter fusion for multilingual Text2Cypher, comparing uniform linear merging and fusion MLP with dynamic adapter weighting. Fusion MLP outperforms uniform linear merging across all three languages.
\item Fusion MLP recovers ~75\% of joint fine-tuning gains using only a subset of the training data (20\% in our experiments), achieving balanced performance across high- and low-resource languages.
\item Fusion MLP supports incremental language expansion, where adding a new language requires only one LoRA adapter and lightweight MLP retraining.
\item We provide a practical framework for multilingual Text2Cypher, combining LLM methodology with graph query generation for real-world database applications.
\end{itemize}

The remainder of the paper is organized as follows: Section~\ref{rel_work} reviews related work, Section~\ref{method} describes our methodology, Section~\ref{results} presents evaluation results, Section~\ref{disc} discusses the findings and Section~\ref{conc} concludes the paper.

\section{Related Work} \label{rel_work}

This section reviews related work on large language models (LLMs) for non-English content, adapter merging and fusion approaches and multilingual methods for database query generation tasks.

\subsection{LLMs and Non-English Content}

Recent advances in LLMs have improved multilingual capabilities~\cite{lai2024llms}. 
Multilingual LLMs are trained on data in multiple languages, enabling knowledge transfer from high-resource languages, such as English, to lower-resource languages. 
However, because most training corpora are English-dominant, these models tend to perform better on English and linguistically similar languages~\cite{nicholas2023lost,zhao2024llama}. 
Even large models exhibit performance gaps for low-resource languages and across different domains~\cite{mishra2025if}, which can impact applications such as natural language interfaces to graph databases.

Several approaches have been proposed to improve multilingual capabilities of LLMs. Some continue pretraining on multilingual parallel corpora~\cite{yang2023bigtranslate,zhu2023extrapolating}, while others fine-tune on multilingual instruction datasets~\cite{ustun2024aya,luo2023yayi,li2023bactrian,lai2023okapi}. Cross-lingual prompting at inference time~\cite{huang2023not,etxaniz2023multilingual} and studies of model internals~\cite{zhao2024llama,kargaran2024mexa,zhong2024beyond,schut2025multilingual,bandarkar2024layer} also aim to improve multilingual performance. While these approaches enhance general language coverage, their applicability to structured query generation tasks such as Text2Cypher remains limited.

\subsection{Adapter Merging and Fusion Methods}

Adapter merging and fusion techniques enable efficient adaptation across tasks and languages without retraining the entire model. Methods such as Task Arithmetic~\cite{ilharco2022editing}, TIES~\cite{yadav2023ties} and DARE~\cite{yu2024language} combine adapters by computing parameter differences and applying scaling or post-processing. LoRA Soups~\cite{prabhakar2025lora} and LoRA-LEGO~\cite{zhao2024merging} present weighted and modular strategies for merging adapters. Dynamic gating approaches like AdapterFusion~\cite{pfeiffer2021adapterfusion} and UniPELT~\cite{mao2022unipelt} select relevant adapters on a per-input basis.

In terms of multilingual support, AdaMergeX~\cite{zhao2025adamergex} separates task and language ability into different adapters and merges them using structure-adaptive merging. MLM~\cite{lee2025mlm} trains a task and a language adapter and combines them through parameter-space interpolation followed by a light post-merging step. 
While these methods have shown promise in multilingual and cross-task transfer, their applicability to structured query generation tasks like Text2Cypher remains unexplored.

\begin{table} [t]
\caption{Instructions used for Text2Cypher task}
  \label{tab:instructions}
  \begin{tabular}{p{0.15\linewidth}p{0.75\linewidth}}
    \hline
    \textbf{Type} & \textbf{Instruction prompt}  \\
     \hline
    System \newline Instruct. &  Task: Generate Cypher statement to query a graph database. Instructions: Use only the provided relationship types and properties in the schema. Do not use any other relationship types or properties that are not provided in the schema. Do not include any explanations or apologies in your responses. Do not respond to any questions that might ask anything else than for you to construct a Cypher statement. Do not include any text except the generated Cypher statement. \\
    \hline
    User \newline Instruct. & Generate Cypher statement to query a graph database. Use only the provided relationship types and properties in the schema. \newline
            Schema: \{schema\} \newline
            \textbf{Question: \{question\} } \newline
            Cypher output: 
         \\
   \hline
\end{tabular}
\end{table}

\subsection{Multilingual Database Query Generation}

Most work on multilingual database query generation from natural language has focused on the Text2SQL task~\cite{dou2023multispider,jose2021mrat,huang2025exploring, pham2025multilingual}. Several studies translate the English Spider dataset~\cite{yu2018spider} into other languages, such as Chinese in CSpider~\cite{min2019pilot}, Turkish in TURSpider~\cite{kanburoglu2024turspider}, Arabic in Ar-Spider~\cite{almohaimeed2024ar} and multiple languages in MultiSpider~\cite{dou2023multispider} and MultiSpider 2.0~\cite{pham2025multilingual}. Other work has created new multilingual datasets, such as StatBot.Swiss~\cite{nooralahzadeh2024statbot} in English and German. Beyond dataset creation, several studies evaluate LLM performance across languages for Text2SQL, for example in Portuguese~\cite{jannuzzi2024zero} and Russian~\cite{bakshandaeva2022pauq}. These studies show that model performance remains uneven across languages, with higher accuracy typically observed for high-resource languages. 
For the Text2SPARQL task, fewer multilingual datasets exist. 
Some work focuses on multilingual question answering over knowledge graphs~\cite{cui-etal-2022-compositional,srivastava2024mst5,perevalov2024multilingual}. Recently, the Text2SPARQL Challenge~\cite{text2sparqlChallenge25} introduced a dataset with questions in English and Spanish. 


\begin{figure*}[t]
\centering
\includegraphics[width=0.8\linewidth]{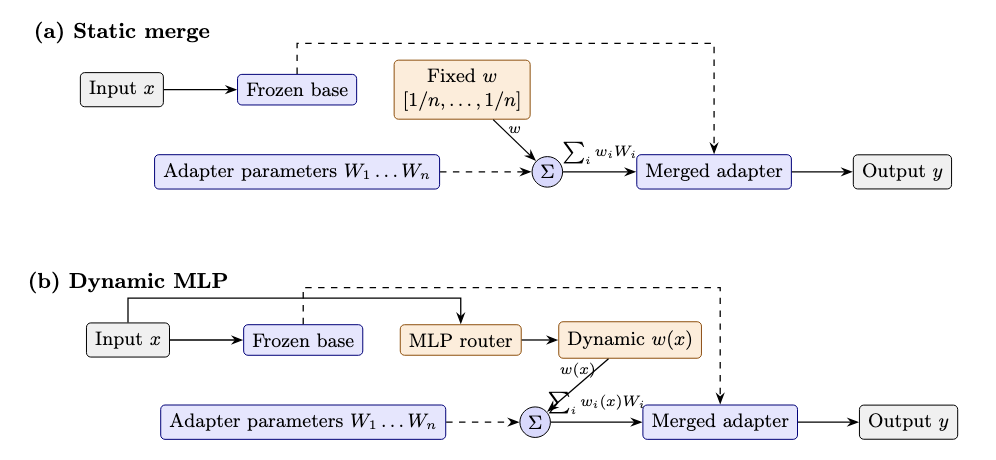}
\caption{At inference-time (a)~Static-linear merge uses fixed weights, (b)~Dynamic MLP computes weights for each input dynamically.}
\label{fig:fusion_methods}
\end{figure*}

For the Text2Cypher task, a multilingual dataset in English, Spanish and Turkish has been created and both base and fine-tuned models are evaluated~\cite{ozsoy2025t2cmultilang}. Their results show that base model performance varies significantly across languages, while joint multilingual fine-tuning reduces these differences. However, full retraining remains computationally expensive for incremental language expansion. 
In this work, we explore adapter fusion to combine language-specific LoRA adapters for multilingual Text2Cypher. This approach enables incremental language expansion while maintaining performance close to joint multilingual fine-tuning.

\section{Methodology} \label{method}

We address challenges in incremental multilingual Text2Cypher by training language-specific LoRA adapters on the benchmark dataset~\cite{ozsoy2025t2cmultilang} and then combining them using either uniform linear merging or a fusion MLP with dynamic adapter weighting. 
Prior work using joint multilingual fine-tuning~\cite{ozsoy2025t2cmultilang} achieves strong performance but requires retraining the full model when new languages are added. 
Figure~\ref{fig:fusion_methods} illustrates how static uniform linear merging and dynamic fusion MLP execute inference.





\subsection{Per-Language Adapters}

For each target language, we train a separate LoRA adapter, for English (EN), Spanish (ES) and Turkish (TR),  on the multilingual Text2Cypher dataset~\cite{ozsoy2025t2cmultilang}, which contains around 12,000 parallel training samples per language. Each sample expresses the same question in different languages, mapping to the same Cypher query. 

LoRA~\cite{hu2022lora} allows efficient fine-tuning by updating a low-rank component of the model's weight matrices while keeping the base model frozen. In our setup, each adapter uses identical hyper-parameters (rank $r=8$, $\alpha=16$) and captures both the Text2Cypher task knowledge and language-specific patterns, such as syntax and vocabulary differences.


These per-language adapters enable incremental language expansion without retraining the entire model. As a result, new languages can be added while maintaining consistent query results across languages. 

\subsection{Uniform Linear Merging}
We combine the per-language LoRA adapters using uniform linear merging~\cite{ilharco2022editing}, assigning each adapter a fixed weight to produce a single merged adapter capable of handling multiple languages. 
In our experiments, we use equal weights for simplicity, 
such that with three adapters corresponding to English, Spanish and Turkish, each is weighted equally ($1/3$) in the merged adapter. This approach provides a simple and efficient baseline without any hyperparameter search.



This approach produces a single multilingual adapter, but the weights are static and do not adapt to differences in input language. It serves as a baseline for comparing with the more flexible fusion MLP method described next.

{\renewcommand{\arraystretch}{1.15}  
\begin{table*}[t]
\centering
\caption{Example dynamic adapter weighting learned by fusion MLP. Spanish and Turkish questions include English translations in parentheses.}
\label{tab:routing_weights}
\begin{tabular}{l|p{0.57\textwidth}|l}
\hline
\textbf{Language} & \textbf{Question} & \textbf{Weights [EN, ES, TR]} \\
\hline
\textbf{English} & Which conferences have the highest number of papers in Robotics presented by authors from 'Tsinghua University' & [\textbf{0.83}, 0.06, 0.11] \\
\textbf{Spanish} & ¿Cuáles son los 5 principales proveedores por número de productos suministrados? \newline \textit{(Which are the 5 main suppliers by number of products supplied?)} & [0.03, \textbf{0.96}, 0.01] \\
\textbf{Turkish} & Satışları en yüksek olan şirketin genel merkezi nerededir? \newline \textit{(Where is the headquarters of the company with the highest sales?)}& [0.07, 0.15, \textbf{0.78}] \\



\hline
\end{tabular}
\end{table*}
}

{\renewcommand{\arraystretch}{1.15}  
\begin{table*}[t]
\centering
\caption{ROUGE-L score on multilingual Text2Cypher test set. Superscript values show absolute improvements over base model.}
\label{tab:rouge_scores}
\begin{tabular}{l|lll|l}
\hline
\textbf{Method} & \textbf{English} & \textbf{Spanish} & \textbf{Turkish} & \textbf{Avg.}\\
\hline
\textbf{Base} & 0.65  & 0.60 & 0.55 & 0.60 \\
\textbf{EN-Only FT} & $0.86 ^{+0.21}$ & $0.80 ^{+0.20}$ & $0.71 ^{+0.16}$ & $0.79 ^{+0.19}$\\
\textbf{Joint Multilang FT} & $0.86 ^{+0.21}$ & $0.85 ^{+0.25}$ & $0.83 ^{+0.28}$ & $0.85 ^{+0.25}$ \\
\hline 
\textbf{Linear Merge} & $0.79 ^{+0.14}$ & $0.76 ^{+0.16}$ & $0.71 ^{+0.16}$ & $0.75 ^{+0.15}$ \\
\textbf{Fusion MLP} & $0.80 ^{+0.15}$ & $0.80 ^{+0.20}$ & $0.78 ^{+0.23}$ & $0.79 ^{+0.19}$ \\
\hline
\end{tabular}
\end{table*} 
}


\subsection{Fusion MLP with Dynamic Adapter Weights}
Unlike static linear merging which uses fixed weights, our fusion MLP predicts input-dependent weights $\mathbf{w}(x)$ that adapt to each input's language characteristics. The complete MLP architecture used for training is detailed in Appendix~\ref{sec:appendix_fusion}.

For each input $x$, we extract adapter preview features from the last 200 tokens of each language-specific adapter's output, capturing language-specific patterns (corresponding to the `Question' section in Table~\ref{tab:instructions}). These features are computed as:

\[
\mathbf{f}_{i_\text{preview}} = \text{Mean}(\text{logits}_i(\text{input\_ids}[:,-200:]))
\]

The preview features from all adapters are concatenated with mean-pooled base model embeddings $h_{\text{base}}$, which are language-agnostic and passed to the MLP:

\[
\mathbf{f}_\text{preview} = [\mathbf{f}_{1_\text{preview}}, \mathbf{f}_{2_\text{preview}}, \dots, \mathbf{f}_{n_\text{preview}}]
\]
\[
\mathbf{w}(x) = \text{softmax}\left(\text{MLP}([\text{Mean}(h_{\text{base}}), \mathbf{f}_\text{preview}])\right)
\]

Here $n=3$ is the number of language adapters (EN, ES, TR) and $\mathbf{w}(x)$ provides dynamic weights (e.g., $[0.75, 0.15, 0.10]$) representing each adapter's contribution for input $x$. 
These dynamic weights are applied differently during training and inference, as described below.

\subsection{Training and Inference}

For training, we first tune the per-language LoRA adapters and then, keeping the the pretrained adapters frozen, we train the fusion MLP using output-level combination. 
The MLP predicts weights $\mathbf{w}(x)$ from preview features (as described above). These weights combine each frozen adapter's full-sequence logits for next-token prediction loss:
\[
\text{logits}_\text{fused} = \sum_{i=1}^n w_i(x) \cdot \text{logits}_i(\text{input\_ids})
\]
where logits are unnormalized vocabulary log-probabilities.


At inference time, the trained fusion MLP predicts dynamic weights $\mathbf{w}(x)$ (See Figure~\ref{fig:fusion_methods}). These weights are used to linearly combine the frozen adapters' LoRA parameters into a single merged module:
\[
W_\text{merged}(x) = \sum_{i=1}^n w_i(x) \cdot W_i
\]
where each $W_i$ denotes adapter $i$'s low-rank update (e.g.\ $W_i = B_i A_i$ in standard LoRA notation), applied to base model hidden states $\mathbf{h}$.
This parameter merging leverages LoRA's linearity~\cite{hu2022lora}.



\section{Experimental Results} \label{results}

We evaluate how merging per-language LoRA adapters via uniform linear merging and learned fusion MLP affects performance on the Text2Cypher task. We compare these modular approaches against the baseline foundational model and a single jointly fine-tuned multilingual model.

\subsection{Experimental Setup}
Experiments use the multilingual Text2Cypher dataset~\cite{ozsoy2025t2cmultilang}, covering English (EN), Spanish (ES) and Turkish (TR). Each sample contains a natural language question, the database schema, the corresponding Cypher query and additional metadata. The test set includes 4,783 parallel samples per language. The training set contains approximately 12,000 samples per language (36,000 total), structured as follows: 6,750 questions shared across all languages, 1,500 questions shared per language pair (EN–ES, EN–TR, ES–TR) and 3,800 questions unique to each language.

We use \texttt{Meta-Llama-3.1-8B-Instruct} as the baseline model. This choice follows prior work~\cite{ozsoy2025t2cmultilang}, enabling direct comparison and reuse of previously generated outputs. For the Text2Cypher task, we adopt the same prompts as prior studies~\cite{ozsoy2025text2cypher, ozsoy2025t2cmultilang} (Table~\ref{tab:instructions}). Generated outputs are post-processed to remove unwanted text, such as the `cypher:' prefix.

Evaluation is performed using both text- and execution-based metrics. Text-based metrics compare the generated and ground-truth Cypher queries. In this work, we report ROUGE-L to quantify their similarity. Execution-based metrics evaluate functional correctness by running the generated and ground-truth queries on the target databases and comparing their outputs. We report Exact-Match, a strict metric that returns a positive score only when the outputs are identical.



{\renewcommand{\arraystretch}{1.15}  
\begin{table*}
\centering
\caption{Comparing training costs when incrementally adding new languages in our Text2Cypher setup.}
\label{tab:scalability}
\begin{tabular}{l|cc}
\hline
\textbf{Languages} & \textbf{Joint Multilang FT} & \textbf{Fusion MLP} \\
\hline
\textbf{1st lang} & 12K instances & 12K instances  \\
\textbf{2nd lang} & 24K instances & 12K instances + (2$\times$2.5K instances) =  17K instances\\
\textbf{3rd lang} & 36K instances & 12K instances + (3$\times$2.5K instances) = 19.5K instances \\
\textbf{4th lang} & 48K instances & 12K instances + (4$\times$2.5K instances) = 22K instances\\
\hline
\end{tabular}
\end{table*}
}

\subsection{Performance Comparison}

We train individual LoRA adapters per language using approximately 12,000 samples each from the multilingual Text2Cypher training set. These adapters are then combined using two strategies detailed in Section~\ref{method}: (1) uniform linear merging and (2) fusion MLP with dynamic adapter weights. 

For uniform linear merging, we assign equal weights ($w_i = 0.333$) to each adapter. This provides a simple, strong baseline without requiring manual hyperparameter tuning. 
For the fusion MLP, we select around 2,500 questions per language from the set of shared questions across all three languages (7,500 total), corresponding to roughly 20\% of the full training data per language. This selection ensures coverage of the 6,750 shared questions while maintaining a balanced per-language representation, allowing the MLP to learn effective routing without using the entire dataset. 

Baselines include the base model, English-only fine-tuning ($\sim$20.5K samples) and joint multilingual fine-tuning ($\sim$36K samples). We replicated prior results~\cite{ozsoy2025t2cmultilang} using the same evaluation setup and obtained comparable metrics. 

Table~\ref{tab:routing_weights} presents example dynamic adapter weighting where the weights are predicted by the fusion MLP for English, Spanish and Turkish questions. Each row lists the input question and the per-adapter weights assigned to English (EN), Spanish (ES) and Turkish (TR) adapters. The fusion MLP correctly assigns the highest weight to the adapter corresponding to the input language, explaining its improved performance over static linear merging. 

Table~\ref{tab:rouge_scores} reports ROUGE-L scores comparing generated and ground-truth Cypher queries. Joint multilingual fine-tuning achieves the highest scores across all languages and serves as an upper bound since it learns from all languages simultaneously. Fusion MLP consistently outperforms uniform linear merging without any hyperparameter search, achieving balanced performance across high- (EN), medium- (ES) and low-resource (TR) languages. It recovers approximately 75\% of the performance gains of joint fine-tuning (0.19 vs 0.25 on average) while using only a fraction of the training data.

{\renewcommand{\arraystretch}{1.15}
\begin{table}
\centering
\caption{Execution-based evaluation on the Text2Cypher test set. Exact-Match metric compare execution outputs of generated queries to ground-truth results.}
\label{tab:exec_scores}
\begin{tabular}{l|lll}
\hline
\textbf{Method} & \textbf{English} & \textbf{Spanish} & \textbf{Turkish} \\
\hline
\textbf{Base} & 0.0973 & 0.0751 & 0.0454 \\
\textbf{Linear Merge} & 0.1303 & 0.1208 & 0.1019 \\
\textbf{Fusion MLP} & 0.1447 & 0.1357 & 0.1072 \\

\hline
\end{tabular}
\end{table}
}

We also computed execution-based Exact-Match scores, presented in Table~\ref{tab:exec_scores}. Linear merging improves Exact-Match by 0.03–0.05, whereas fusion MLP yields slightly higher gains of 0.05–0.06. Although Exact-Match is a very strict metric, the observed improvements align with ROUGE-L trends, providing additional evidence that adapter fusion enhances functional consistency across languages.

\subsection{Training Efficiency}

A key advantage of the fusion MLP approach is its ability to scale incrementally when adding new languages. In contrast, joint multilingual fine-tuning requires retraining the full model whenever a new language is introduced (e.g., increasing from 24K to 36K to 48K training samples). Fusion MLP avoids this overhead, such that each new language requires only (1) training a separate LoRA adapter (e.g., $\sim$12K samples) and (2) updating the fusion MLP (e.g., $\sim$2.5K samples), while previously trained adapters remain frozen.

Table~\ref{tab:scalability} illustrates this efficiency using a hypothetical scenario in our Text2Cypher setup. For instance, adding a fourth language would require 48K samples for joint fine-tuning but only 22K samples for the fusion MLP approach, representing a 54\% reduction in training data. This modular design enables continuous language expansion without without the computational cost of full retraining.



\section{Discussion} \label{disc}

Fusion MLP consistently outperforms linear merging, achieving an average ROUGE-L of 0.79 versus 0.75. For the low-resource language Turkish, fusion MLP reaches 0.78, improving over the base model (0.55) and English-only fine-tuning (0.71), though joint multilingual fine-tuning still achieves the highest score (0.83). These results demonstrate that adapter fusion effectively leverages knowledge from high- and medium-resource languages to improve query accuracy in languages with limited training data, ensuring reliable results without full model retraining.

While fusion MLP enables incremental language expansion and balanced performance across languages, several limitations remain. Our evaluation covers only English, Spanish and Turkish and broader testing across more languages, domains and graph schemas is needed. We focus on uniform linear merging and fusion MLP, leaving other adapter fusion methods unexplored. Inference-time overhead and design choices, such as the fixed 200-token preview window, could also be further optimized.

Our experiments were scoped to highlight the core contribution of adapter fusion and incremental scaling, particularly for low-resource languages. This allows reproducibility and direct comparison to prior work, without requiring exhaustive evaluation across all model sizes, datasets, or fusion strategies.

\section{Conclusion} \label{conc}

This work proposes a practical framework for multilingual Text2Cypher that addresses incremental language expansion, automated hyperparameter management and competitive performance without full model retraining. We train language-specific LoRA adapters for English, Spanish and Turkish and combine them using either uniform linear merging or fusion MLP with dynamic adapter weighting.

Fusion MLP outperforms linear merging across all languages and achieves balanced performance, especially benefiting low-resource languages such as Turkish. It recovers approximately 75\% of the gains from joint multilingual fine-tuning using only a subset of training data. The modular design allows adding new languages by training a single LoRA adapter and updating the lightweight MLP, without retraining existing adapters.

Overall, learned adapter fusion provides a practical and efficient alternative to full multilingual fine-tuning. It delivers balanced performance across high-, medium- and low-resource languages and supports continuous language expansion. Future work will extend this approach to additional languages and explore alternative fusion strategies to further improve multilingual Text2Cypher performance.


  

\section*{Declaration on Generative AI Usage}

During the preparation of this work, the author(s) used Generative AI tools in order to: 'Improve writing style', 'Paraphrase and reword', 'Code debugging and fixes'. 
After using these tool(s) or service(s), the author(s) reviewed and edited the content as needed and take(s) full responsibility for the publication's content.





\bibliographystyle{ACM-Reference-Format}
\bibliography{main}

\appendix
\section*{Appendix}

\section{Fine-tuning parameters}
For fine-tuning, we used a RunPod \cite{runpod} GPU environment with a single A40 machine. 
The parameters used are presented in Table \ref{tab:finetune_parameters}. 


\begin{table}[H]
  \caption{Fine-tuning Parameters}
  \label{tab:finetune_parameters}
  \begin{tabular}{p{0.22\linewidth}|p{0.70\linewidth}}
    \hline
     \textbf{Model \& \newline Tokenizer \newline Parameters} & 
        max\_seq\_length: 2048 \newline
        dtype: torch.bfloat16 \newline
        load\_in\_4bit: False \newline
        truncation\_side: "left" \newline
        padding\_side: "left"
    \\
    \hline
    \textbf{PEFT \newline Parameters} &  
        r: 8 \newline
        target\_modules: \newline [
            "q\_proj",
            "k\_proj", \newline
            "v\_proj",
            "o\_proj"
        ] \newline
        lora\_alpha: 16 \newline
        lora\_dropout: 0 
     \\
    \hline
    \textbf{Training \newline Arguments} &   
        per\_device\_train\_batch\_size: 2 \newline
        gradient\_accumulation\_steps: 4\newline
        warmup\_steps: 5 \newline
        num\_train\_epochs: 1 \newline
        learning\_rate: 2e-4 \newline
        optim: "adamw\_8bit" \newline
        weight\_decay: 0.01 \newline
        lr\_scheduler\_type: "linear" 
     \\
  \hline
\end{tabular}
\end{table}

\section{Fusion MLP Architecture}\label{sec:appendix_fusion}

\begin{figure}[H]
  \centering
  \includegraphics[width=0.94\linewidth]{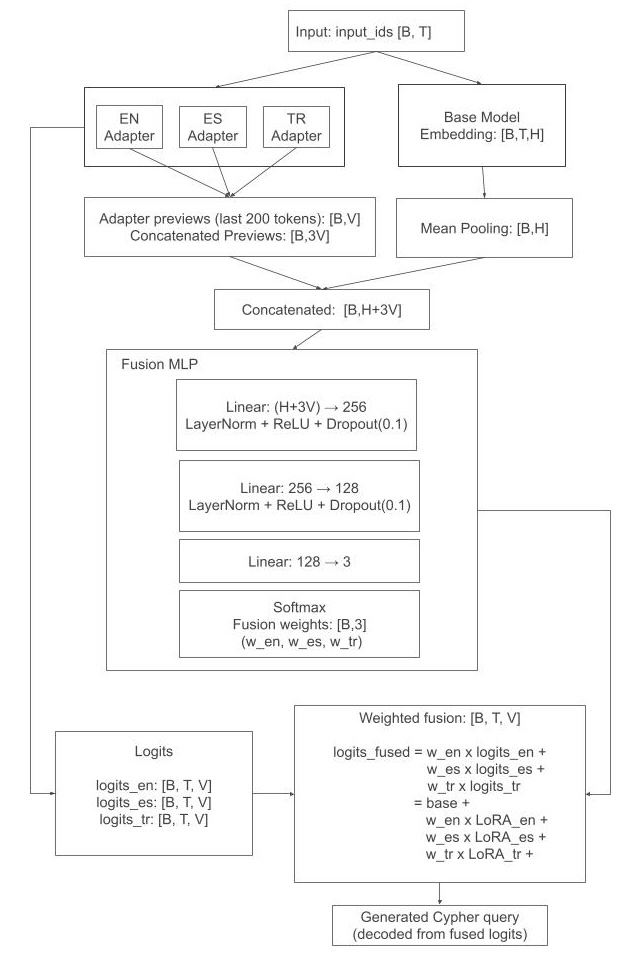}
  \caption{Architecture of fusion MLP with dynamic adapter weighting.}
  \label{fig:fusion_arch}
\end{figure}

The fusion MLP architecture for training is shown in Figure~\ref{fig:fusion_arch}.
Notation: $B$ = batch size, $T$ = sequence length, $H$ = hidden dimension, $V$ = vocabulary size, $3$ = number of adapters (EN, ES, TR).

\end{document}